\documentclass[10pt,twocolumn,letterpaper]{article}
\usepackage{url}
\usepackage{cvpr}
\usepackage{times}
\usepackage{epsfig}
\usepackage{graphicx}
\usepackage{amsmath}
\usepackage{amssymb}

\usepackage{booktabs}
\usepackage{multirow}
\usepackage{subfig}

\usepackage[breaklinks=true,bookmarks=false,colorlinks]{hyperref}

\cvprfinalcopy 

\def\cvprPaperID{****} 

\begin{document}

\title{1st Place Solutions for Waymo Open Dataset Challenges - 2D and 3D Tracking}

\author{Yu Wang \quad Sijia Chen \quad Li Huang \quad Runzhou Ge \\ Yihan Hu  \quad Zhuangzhuang Ding \quad Jie Liao\\
[1.0ex]
Horizon Robotics Inc.\\
{\tt\small yuwangrpi@gmail.com}
}

\maketitle

\begin{abstract}
  This technical report presents the online and real-time 2D and 3D multi-object tracking (MOT) algorithms that reached the 1st places on both Waymo Open Dataset 2D tracking and 3D tracking challenges. An efficient and pragmatic online tracking-by-detection framework named HorizonMOT is proposed for camera-based 2D tracking in the image space and LiDAR-based 3D tracking in the 3D world space. Within the tracking-by-detection paradigm, our trackers leverage our high-performing detectors used in the 2D/3D detection challenges and achieved $45.13\%$ 2D MOTA$/$L2 and $63.45\%$ 3D MOTA$/$L2 in the 2D/3D tracking challenges.
\end{abstract}

\begin{figure}
\centering
\subfloat[]{
\includegraphics[width=0.9\linewidth]{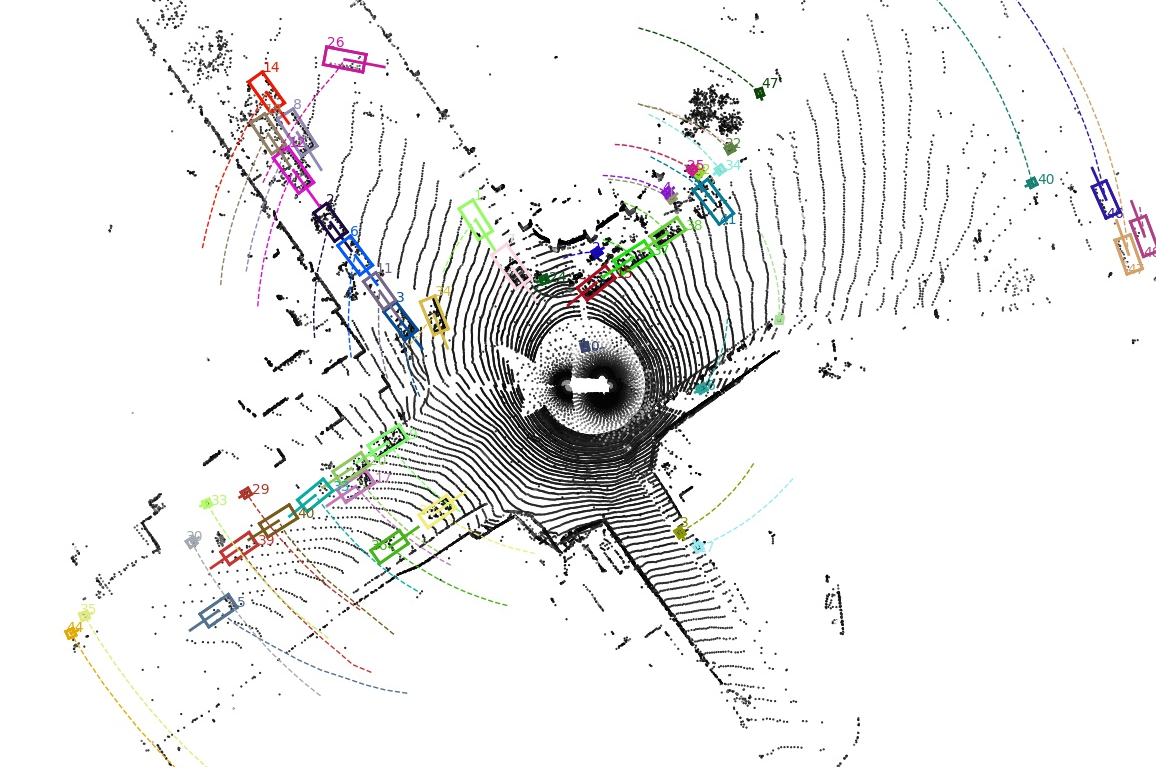}
 \label{fig:subfig1}
}\\
\subfloat[]{
\includegraphics[width=1.0\linewidth]{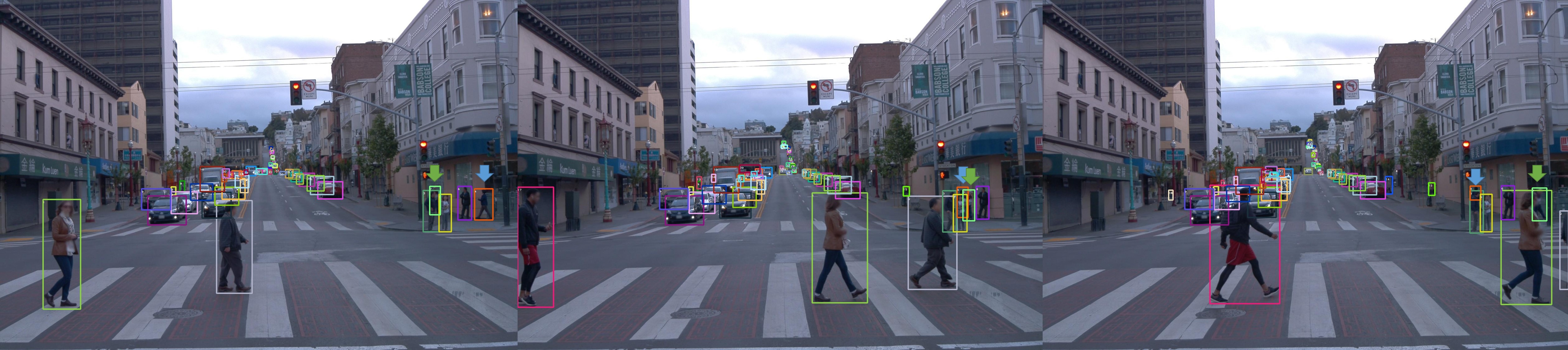}
\label{fig:subfig2}
}\\
\subfloat[]{
\includegraphics[width=1.0\linewidth]{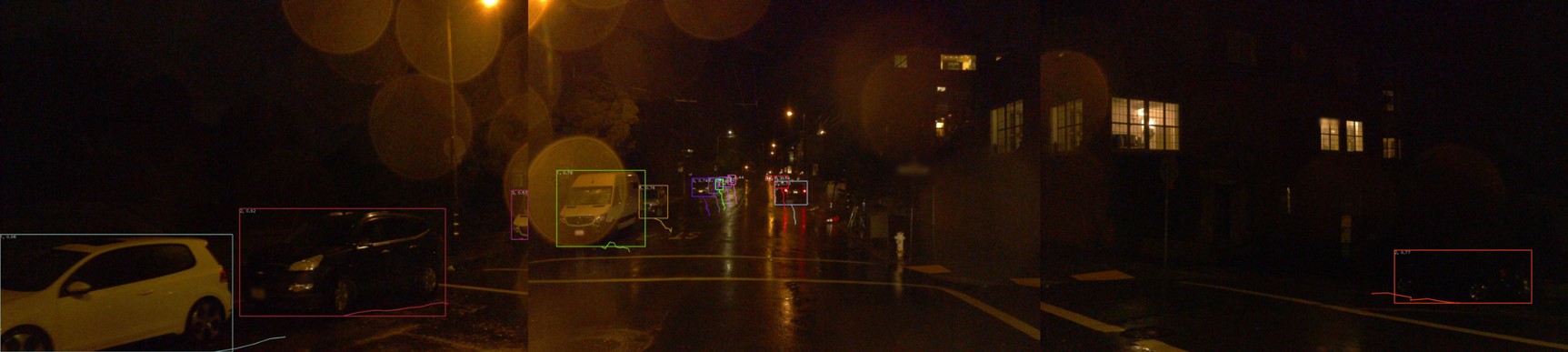}
 \label{fig:subfig3}
}
\caption{Examples of our 3D tracking (a) and 2D tracking (b)(c) results on the Waymo Open Dataset v1.2. The ego-vehicle is making a left turn in (a). We highlight in (b) two pedestrians that were tracked successfully even after occlusions in a crowed scene. Note in (c) that we do not track the object across cameras as it is not required by the challenge. Trajectories of the cars in the front camera (c) indicate large object displacement caused by pitch motion of the camera possibly due to uneven ground.}
\label{fig:main_illustrations} 
\end{figure}

\section{Introduction}
Tracking-by-detection approaches have been leading the online (no peeking into the future) multi-object tracking (MOT) benchmarks, as a result of the high-performing object detection models. There are a group of pragmatic tracking-by-detection approaches for 2D/3D multiple object tracking whose data association method is simply based on bounding box overlap or object center distance and built upon Kalman filter~\cite{Wojke2017simple}~\cite{Bewley_2016}~\cite{1517Bochinski2017}~\cite{weng2019baseline}~\cite{chiu2020probabilistic}. The majority of the participants in Waymo 2D and 3D tracking challenges are based on these methods. Despite the fact that tracking-by-detection often relies on strong object detectors, better overall performance can still be achieved by improving the data association schemes and the tracking framework. 

In recent literature, there is a trend of joint detection and tracking using a single network, such as RetinaTrack~\cite{lu2020retinatrack}, and CenterTrack~\cite{zhou2020tracking} and FairMOT~\cite{zhang2020simple} which are developed on top of CenterNet. This paradigm was adopted by some participants in the 2D tracking challenge. The CenterTrack ~\cite{zhou2020tracking} network learns a 2D offset of the same object between two adjacent frames and associate it over time based on center distance. The overall idea of CenterTrack is simple yet effective. One problem of CenterTrack is that it achieves tracking using the center offsets in the local regime and therefore is unable to handle the long-term occlusion or missing detection problems. 

For the 2D and 3D tracking challenges, we proposed a unified and pragmatic framework named HorizonMOT that focuses on frame-to-frame prediction and association, and is applicable to both 2D camera-based tracking in the image space and LiDAR-based 3D tracking in the 3D world space, as shown in Figure~\ref{fig:main_illustrations}. Our trackers are online since only detections of the current frame are presented to the tracker, and the result of the current frame is decided right away without any latency. Our trackers belong to the tracking-by-detection paradigm. 

\section{Detection Network}
\subsection{2D Detection Network} High-performing detectors are the key to the success of tracking-by-detection approaches. We employ the one-stage, anchor-free, non-maxima suppression (NMS) free CenterNet framework~\cite{zhou2019objects} for 2D object detection. Under the CenterNet paradigm, many complicated perception tasks can be simplified in a unified framework as object center point detection and regression of object properties such as bounding box size, 3D information (\eg 3D location, 3D dimension, heading), pose, or embedding. 

\begin{figure}
\includegraphics[width=0.9\linewidth]{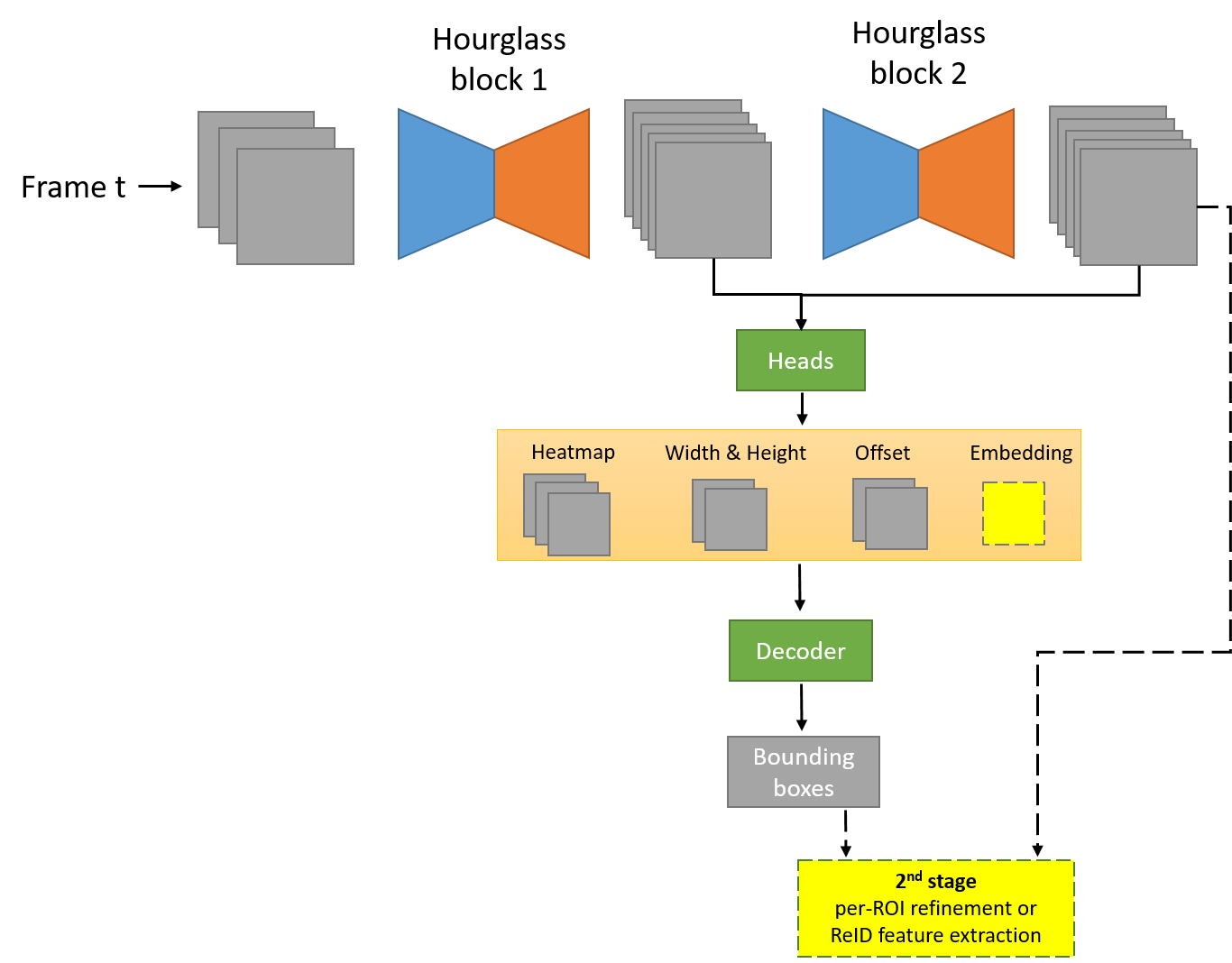}
\caption{The main detection network is based on CenterNet with Hourglass-104 as the backbone. We did not finish the embedding head and the 2nd-stage per-ROI refinement or Re-ID feature extraction branches before the challenge deadline and therefore leave them as future work.}
\label{fig:main_network} 
\end{figure}

We use Hourglass as the CenterNet backbone. As illustrated in Figure~\ref{fig:main_network}, two hourglass blocks are stacked and the first one only serves as providing auxiliary loss during training. We tried using both stacks for inference but it did not improve the results. 

For a more unified and versatile network, we can add a pixel-wise embedding (similar to ~\cite{zhang2020simple}) and a second-stage per-ROI feature extraction branch to extact Re-ID features, which however were not used for this challenge and we leave this for future work.

\subsection{3D Detection Network}
In the 3D detection track of Waymo Open Dataset Challenges, our solution is an improvement upon our baseline 3D point cloud detector named AFDet~\cite{ge2020afdet} and reached the 1st place, and we use 3D detections produced by this solution as input to our 3D tracker.

\section{Tracking Framework}
\begin{figure}
\centering
\includegraphics[width=1.0\linewidth]{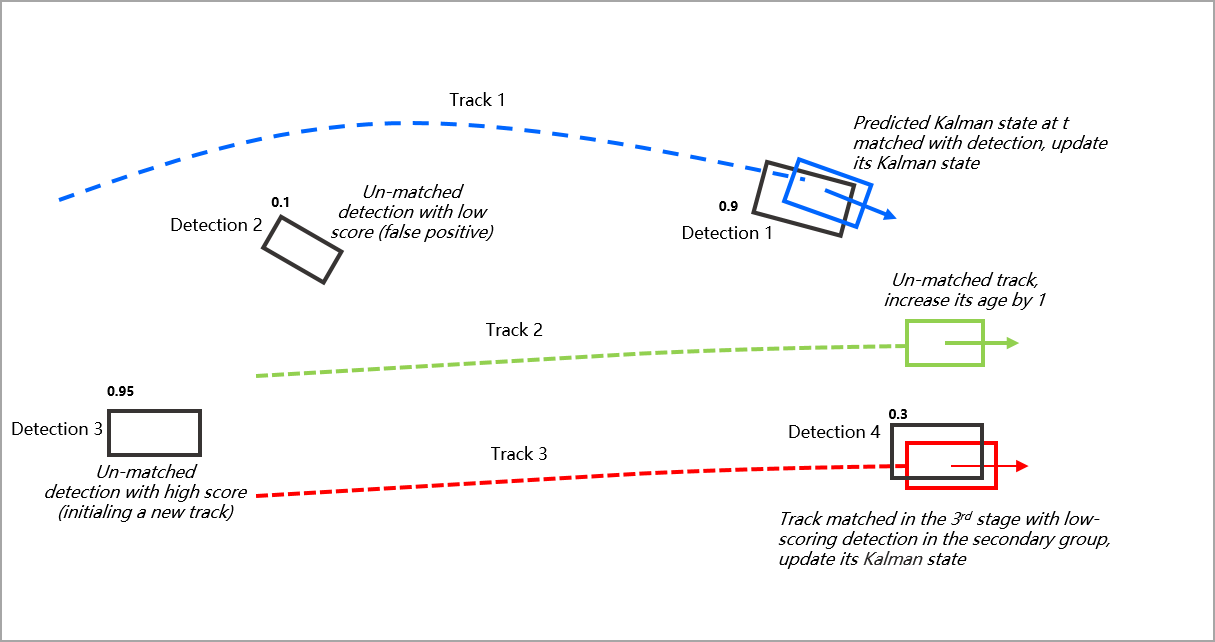}
\caption{An illustration of the tracking framework and data association. Track 1 and 3 are associated with detection 1 and 4 respectively and therefore updated, while track 2 is unmatched and will be deleted if its age exceeds the specified threshold. Detection 2 is considered as false positive and detection 3 is used to initiate a new track.}
\label{fig:tracking-framework} 
\end{figure}

Tracking-by-detection consists of the following components: 1) track creation and deletion; 2) state prediction and update using the Kalman filter; 3) association between tracks and detections. We assume no ego-motion information and no future information is available. An illustration of our tracking framework and data association method is shown in Figure~\ref{fig:tracking-framework}.

\subsection{Track Creation and Deletion}
Similar to~\cite{Wojke2017simple} and~\cite{weng2019baseline}, a new track is created when a detection of the current frame is not associated with any track. As in~\cite{Wojke2017simple} and~\cite{weng2019baseline}, each track $k$ has a number of frames since the last successful detection association $a_{k}$ and the track will be deleted once this counter exceeds predefined maximum age $A_{max}$.

\subsection{State Prediction and Update Using Kalman Filter}
In 2D tracking, for each track we define an eight dimensional state space $(x, y, \gamma, h, \dot{x}, \dot{y}, \dot{\gamma}, \dot{h})$, which contains the center $(x,y)$, aspect ratio $\gamma$, height ${h}$, and their respective velocities in the image space. The observation is 2D detection box and its score $(x, y, w, h, s)$. We simply set the score of the track to the score of its associated detection. In 3D tracking we use 10-dimensional state space $(x, y, z, h, w, l, \theta, \dot{x}, \dot{y}, \dot{z})$ which contains the 3D location, height, width and length, heading, and the respective velocities of the 3D location values, in the 3D space. The observation is 3D detection defined as $(x, y, z, h, w, l, \theta, s)$ where $s$ is the detection score. At each frame, state prediction is performed first using a constant velocity model, and then the track-detection association. The state of each track is updated if it is associated with a detection. In Kalman filter, the estimated 2D/3D box is essentially a weighted average between state space and the observation ~\cite{weng2019baseline}. In our experiments we use the observation directly as the output instead of using the weighted average. If a track is not associated with any detections at the current frame, only the prediction step is performed and the track does not contribute to the output of the current frame.

\subsection{{Association Metric}}
\begin{figure}
\centering
\includegraphics[width=0.9\linewidth]{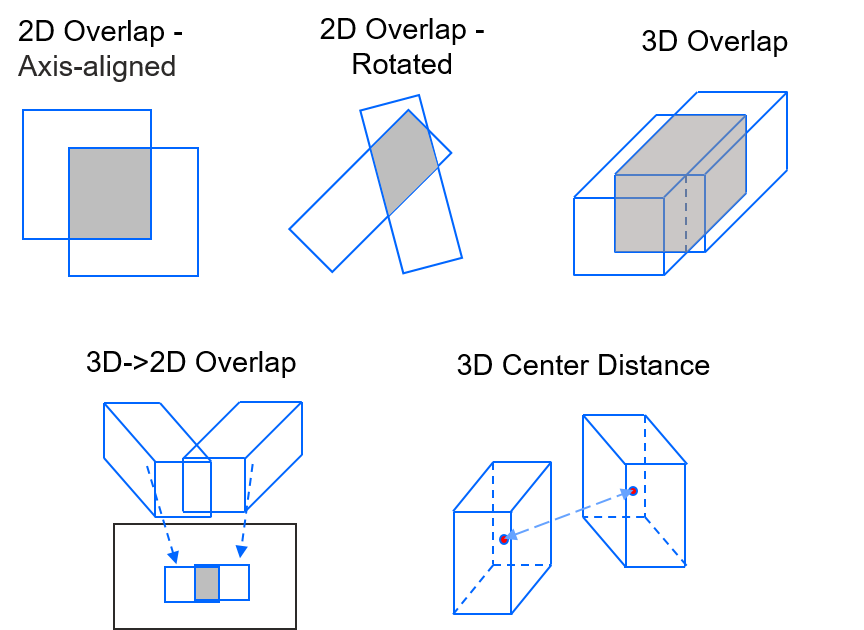}
\caption{A group of association metrics based on 2D/3D bounding boxes.}
\label{fig:bounding-boxes-distance} 
\end{figure}

The association between detections of the current frame and tracks is based on association metrics which are usually defined based on 2D/3D IoUs~\cite{Bewley_2016}~\cite{1517Bochinski2017}~\cite{weng2019baseline}, Mahalanobis distance of 2D/3D object centers~\cite{Wojke2017simple}~\cite{chiu2020probabilistic}, and cosine distance~\cite{Wojke2017simple} between appearance/Re-ID features of 2D boxes. For 3D tracking in the 3D world space, one can also transform the 3D bounding boxes to image space and calculate association metric based on the overlap of 2D projections as shown in Figure~\ref{fig:bounding-boxes-distance}. In the 2D tracking challenge, we adopt 2D IoU and cosine distance, while in the 3D tracking challenge we employ Euclidean distance with Gaussian kernel (with a parameter $\sigma$) between 3D centers, which is better and faster than other metrics that we tried on the validation set such as 3D box IoU, Bird's Eye View (BEV) box IoU (\ie ignoring the vertical dimension) and Mahalanobis distance.

\subsection{Three-stage Data Association}
Typically, association between tracks and detections is formulated as an assignment problem and relies on the Hungarian algorithm. In the tracking challenges, we developed a three-stage data association scheme that applies to both 2D and 3D tracking to improve the tracking performance. We first select a primary set of detections whose scores are larger than $t^{(s)}$ and a secondary set in which scores are within the range $[t^{(s)}/2, t^{(s)})$. 

\textbf{First-stage Association}. We adopt the matching cascade proposed in~\cite{Wojke2017simple} for the first stage. Association cost matrix is calculated first between tracks and the primary set of detections. We exclude unlikely associations if the cost is larger than a specified threshold $t^{(1)}$. We start from the most frequently seen track (\ie with smallest track age $a_{k}$) and iterate over each track of increasing age and solve a linear assignment problem. 

\textbf{Second-stage Association}. In the second stage, the association is between un-matched tracks with age less than 3 and remaining detections in the primary detection set, we use a different association metric or relax the condition of the same association metric used in the first stage (\eg by enlarging the size of a 2D bounding box to increase its overlap over time). The association is again solved in a linear assignment problem and only admissible associations are kept by excluding unlikely associations using a specified distance threshold $t^{(2)}$.

\textbf{Third-stage Association}. In the third matching stage, association is between remaining un-matched tracks and detections in the secondary set. This helps to account for objects with weak detections (\eg caused by partial occlusion). Admissible associations with distance lower than specified threshold $t^{(3)}$ are kept.

\subsection{Re-ID Features}
Our 2D tracker also relies on Re-ID features extracted by a small independent network to complement bounding box overlap based association metrics. Re-ID or appearance features help handle long-term occlusion or objects with large displacement which could result in the failure of IoU based association metric. There are many scenarios which could lead to rapid displacements of object in the image plane. For example, low frame rate, vehicles in the opposite traffic direction with high relative speed, and unaccounted camera motion such as large camera pitch motion caused by bumps on the ground.

Following ~\cite{Wojke2017simple}, we keep a gallery of the history associated Re-ID features of each track and the smallest cosine distance between them and the detection is used as the distance. We also introduce a maximum appearance distance $t^{(a)}$ to exclude unlikely associations.  

\begin{table}[htp]
\setlength{\tabcolsep}{1pt}
\begin{center}
\begin{tabular}{l|ccc}
  \toprule
  \multirow{2}{*}
  \textbf{2D Tracking Parameters} & \textbf{Pedestrian} & \textbf{Vehicle} & \textbf{Cyclist}\\
    \midrule
    \midrule
  score threshold $t^{(s)}$  & 0.5 & 0.4 & 0.5  \\ 
  max\_appearance\_dist $t^{(a)}$ & 0.15 & 0.06 & 0.15 \\
  max\_iou\_dist (front) & 0.95 & 0.9 & 0.95 \\
  max\_iou\_dist (front left/right) & 0.97 & 0.93 & 0.97 \\
  max\_iou\_dist (side) & 0.99 & 0.95 & 0.99 \\
    \bottomrule
\end{tabular}
\end{center}
\caption{2D tracking parameters. Maximum appearance distance and IoU distance are used to exclude unlikely associations during the three-stage data association.}
\label{tab:2d_tracking_parameters}
\end{table}

\begin{table}[htp]
\setlength{\tabcolsep}{2pt}
\begin{center}
\begin{tabular}{l|ccc}
  \toprule
  \multirow{2}{*}
  \textbf{3D Tracking Parameters} & \textbf{Pedestrian} & \textbf{Vehicle} & \textbf{Cyclist}\\
    \midrule
    \midrule
  score threshold $t^{(s)}$  & 0.5 & 0.5 & 0.5  \\ 
  max\_center\_dist & 0.7 & 0.5 & 0.9 \\
  Gaussian kernel $\sigma$ & 1.5 & 5 & 3 \\
    \bottomrule
\end{tabular}
\end{center}
\caption{3D tracking parameters. Maximum center distance is used to exclude associations with unlikely displacement in the 3D world space.}
\label{tab:3d_tracking_parameters}
\end{table}

\section{Experiments}
\subsection{Settings}

\textbf{Dataset}.
Our tracking algorithms are evaluated on the Waymo Open Dataset v1.2~\cite{sun2019scalability}. We use its training set for training the 2D detection networks and 2D Re-ID networks, its validation set for verifying ideas and tuning parameters, and the test set to generate our final submission to the leaderboard~\cite{WaymoChallenge2020}.

\begin{table*}[t!]
\setlength{\tabcolsep}{2pt}
\centering
\begin{tabular}{lccccccc}
\toprule
 & \textbf{MOTA$/$L1} $\uparrow$ & \textbf{MOTP$/$L1} $\uparrow$ & \textbf{MOTA$/$L2} $\uparrow$ & \textbf{MOTP$/$L2} $\uparrow$ & \textbf{FP$/$L2} $\downarrow$ & \textbf{Mis-match$/$L2} $\downarrow$ & \textbf{Miss$/$L2} $\downarrow$   \\
\midrule
\midrule
HorizonMOT & 51.01 & 14.18 & 45.13 & 14.27 & 7.13 & 2.25 & 45.49 \\
Quasi-Dense R101 & 51.18 & 15.12 & 45.09 & 15.20 & 7.20 & 1.31 & 46.41 \\
CascadeRCNN-SORTv2 & 50.22 & 14.85 & 44.15 & 14.85 & 6.94 & 2.44 & 46.46 \\
Online V-IOU & 46.07 & 13.28 & 40.09 & 13.40 & 5.73 & 3.42 & 50.76 \\
DSTNet & 43.83 & 15.76 & 38.01 & 15.84 & 6.28 & 1.34 & 54.38 \\
\bottomrule
\end{tabular}
    \caption{Top-5 2D tracking results on the test set of Waymo Open Dataset, with MOTA/L2 as the primary metric.}
\label{tab:2d_mot_results}
\end{table*}

\begin{table*}[t!]
\setlength{\tabcolsep}{4pt}
\centering
\begin{tabular}{lccccccc}
    \toprule
     & \textbf{MOTA$/$L1} $\uparrow$ & \textbf{MOTP$/$L1} $\uparrow$ & \textbf{MOTA$/$L2} $\uparrow$ & \textbf{MOTP$/$L2} $\uparrow$ & \textbf{FP$/$L2} $\downarrow$ & \textbf{Mis-match$/$L2} $\downarrow$ & \textbf{Miss$/$L2} $\downarrow$   \\
    \midrule
    \midrule
HorizonMOT3D & 65.13 & 23.96 & 63.45 & 23.96 & 7.28 & 0.29 & 28.99 \\
PV-RCNN-KF & 57.14 & 24.95 & 55.53 & 24.97 & 8.66 & 0.63 & 35.18 \\
Probabilistic & 49.16 & 24.80 & 47.65 & 24.82 & 8.99 & 1.01 & 42.35 \\
3DMOT FS-H & 46.52 & 24.69 & 45.07 & 24.74 & 9.04 & 1.90 & 44.00 \\
3DMOT-MD-TPF & 44.03 & 26.30 & 42.56 & 26.31 & 10.07 & 0.44 & 46.93 \\
    \bottomrule
\end{tabular}
    \caption{Top-5 3D tracking results on the test set of Waymo Open Dataset, with MOTA/L2 as the primary metric.}
\label{tab:3d_mot_results}
\end{table*}

\textbf{Evaluation Metric}.
Waymo Open Dataset uses multiple object tracking metrics from ~\cite{MOT2008}. MOTA is the main metric that takes into account the number of misses, false positives and mismatches. It is calculated for two difficulty levels. L1 metrics are calculated only for level 1 ground truth, while L2 metrics are computed by considering both level 1 and level 2 ground truth. 

\subsection{Implementation Details}
\textbf{2D and 3D Detections}. In contrast to the original CenterNet, we use Gaussian kernels as in~\cite{liu2019trainingtimefriendly} which takes into account the aspect ratio of the bounding box to encode training samples. During training, we use $768\times1152$ as the input size and a learning rate of $1.25e$-4. Due to lack of computational resource and sheer size of the dataset, we first trained a main network with weights pretrained on COCO on a $1/10$ subset of the training images for all 3 object categories (\ie car, pedestrian, cyclist) for 25 epochs. A daytime expert model and a nighttime expert model were fine-tuned from this main network using only daytime or nighttime training images in the subset. To handle the imbalanced training data problem (\ie pedestrian and especially cyclist have significantly less training samples than vehicle class), we also fine-tuned an expert model using only images with pedestrian and cyclist samples. We then fine-tuned 4 more models on the entire validation set, the entire training set, and images in the entire training set with pedestrian and cyclist samples, and nighttime images in the entire training set, respectively for 8-10 epochs. In inference we use flip and multi-scale (0.5, 0.75, 1, 1.25, 1.5) augmentation. To serve as the tracker input, outputs of the 8 models were merged by naive NMS with IoU overlap threshold set to 0.5. Note that in the 2D detection challenge we use weighted boxes fusion instead to merge the results.

As input to our 3D tracker, we rely on the 3D detections produced by our solution in the 3D detection challenge. Details of this solution can be found in our technical report for that challenge.

\textbf{Re-ID Network}.
We use an independent Re-ID network with 11 $3\times3$ and 3 $1\times1$ convolutional layers and a maxpooling and average pooling layer and a downsampling factor of 16. Input image is normalized to $128\times64$ for pedestrian, and $128\times128$ for car/cyclist. The network was trained from scratch as a classification network by adding a fully-connected layer and we prepared a total of 2844, 20041, and 906 unique objects for the pedestrian, car, and cyclist respectively from a subset of the Waymo 2D training images. The classification layer is removed during inference and the 512-dimension feature embedding servers as Re-ID feature.

\textbf{2D Tracking}.
Cosine distance between Re-ID features is used in the first-stage matching, 2D IoU distance is used in the second and third-stage matching. We double or triple the size of the bounding boxes in the second and third-stage respectively when calculating the IoU overlap to account for objects with large displacement. Table~\ref{tab:2d_tracking_parameters} summarizes all the parameters used in my 2D tracking experiments. Note that we use different IoU matching thresholds for front, front left and right, and side cameras. We allow larger IoU distance (\ie smaller overlap) in admissible associations for front left/right and side cameras since the displacement of some objects (especially pedestrians) tend to be very large. We assign the score of associated detection to the track as its score.

\textbf{3D Tracking}.
Euclidean distance with Gaussian kernel between 3D centers is used throughout the three-stage associations. We use different $\sigma$ values for each class as shown in Table~\ref{tab:3d_tracking_parameters}. We also assign the score of associated detection to the track as its score.

\begin{table}[htp]
\begin{center}
\begin{tabular}{l|cc}
  \toprule
  \multirow{2}{*}
  \textbf{Module Name} & \textbf{MOTA$/$L1} & \textbf{MOTA$/$L2}\\
    \midrule
    \midrule
  Baseline  & 41.73 & 36.03 \\ 
  + Third-stage association & 46.10 & 39.68 \\ 
  + Re-ID models & 48.79 & 42.11 \\
    \bottomrule
\end{tabular}
\end{center}
\caption{Ablation study on the 2D tracking validation set}
\label{tab:2d_ablation_study}
\end{table}

\subsection{Results}
\begin{figure*}
\centering
\subfloat{
\includegraphics[width=0.9\linewidth]{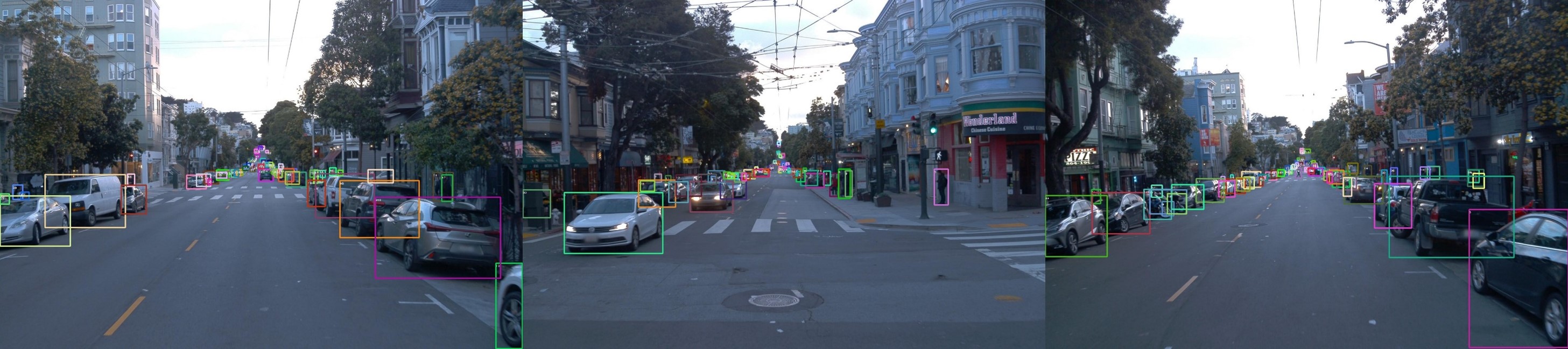}  
 \label{fig:subfig1}
}\\
\subfloat{
\includegraphics[width=0.9\linewidth]{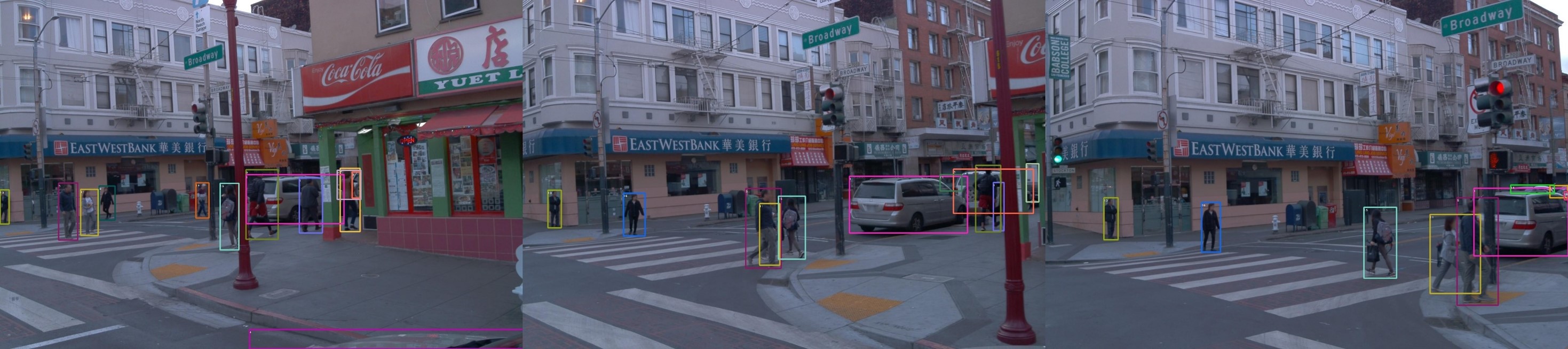}  
\label{fig:subfig2}
}\\
\subfloat{
\includegraphics[width=0.9\linewidth]{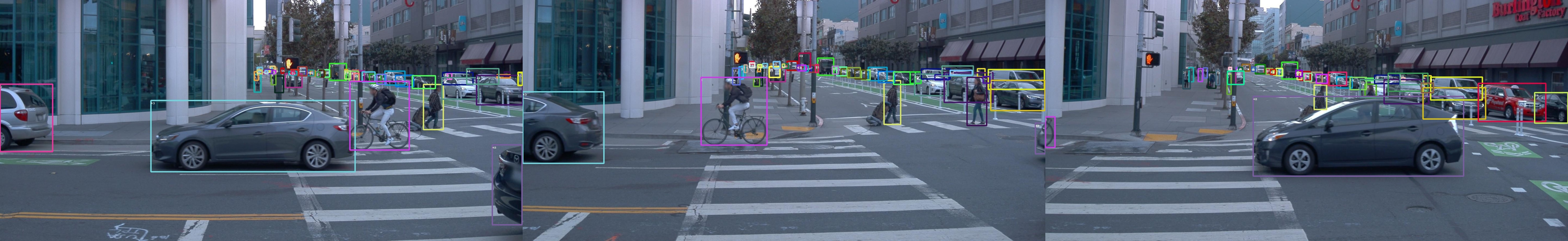}  
\label{fig:subfig3}
}\\
\subfloat{
\includegraphics[width=0.9\linewidth]{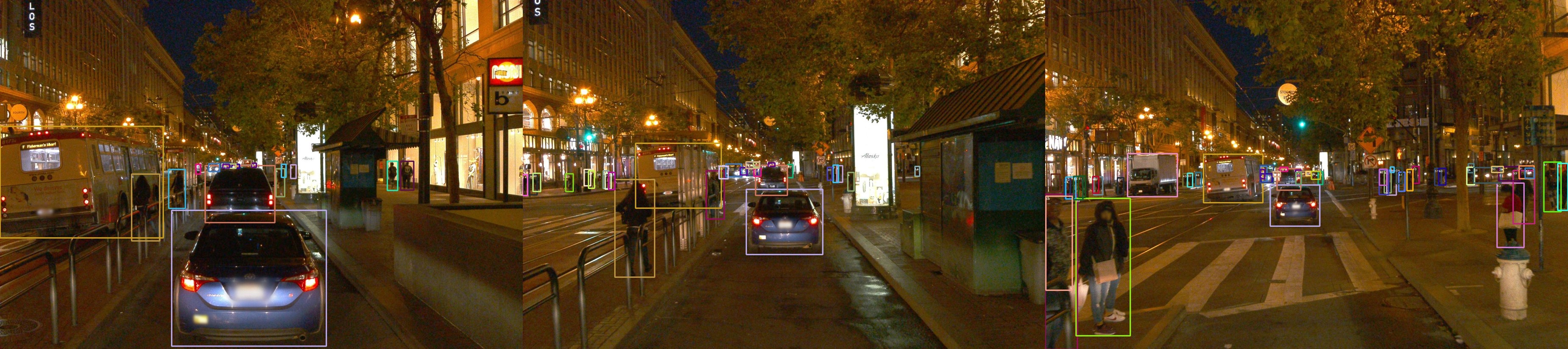}  
 \label{fig:subfig4}
}\\
\subfloat{
\includegraphics[width=0.9\linewidth]{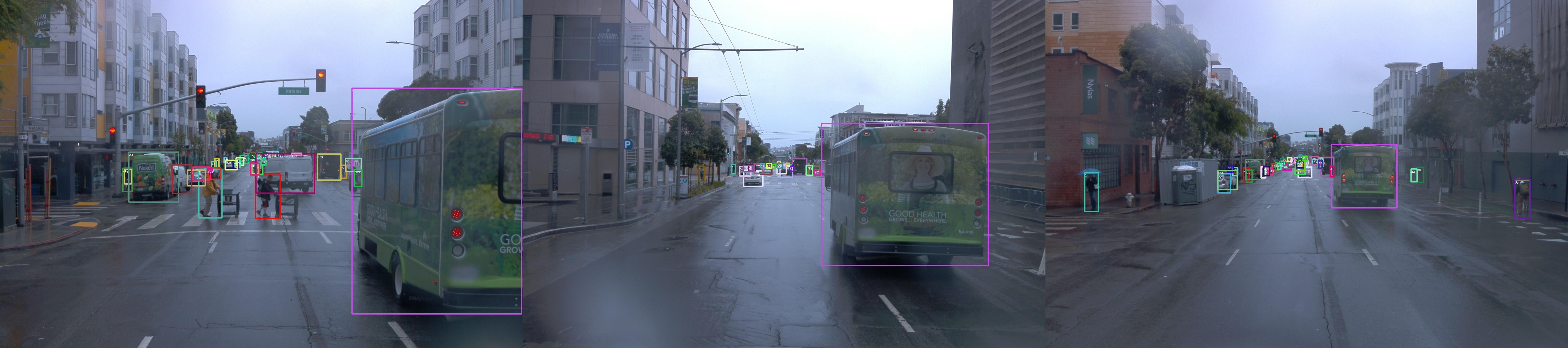}  
 \label{fig:subfig4}
}
\caption{Qualitative results of 2D tracking. Each row shows 3 frames from the same sequence. Each object is showed in a unique color and assigned a unique ID. Note that they are not consecutive frames, the time interval is larger than one frame.}
\label{fig:2d_tracking_qualitative_results} 
\end{figure*}

\begin{figure*}
\centering
\subfloat{
\includegraphics[width=0.9\linewidth]{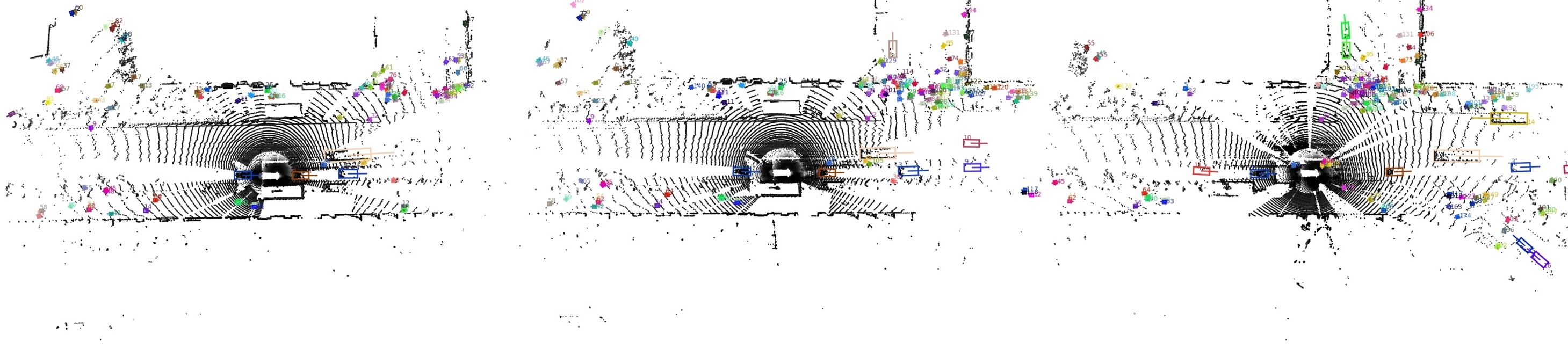}  
 \label{fig:subfig1}
}\\
\subfloat{
\includegraphics[width=0.9\linewidth]{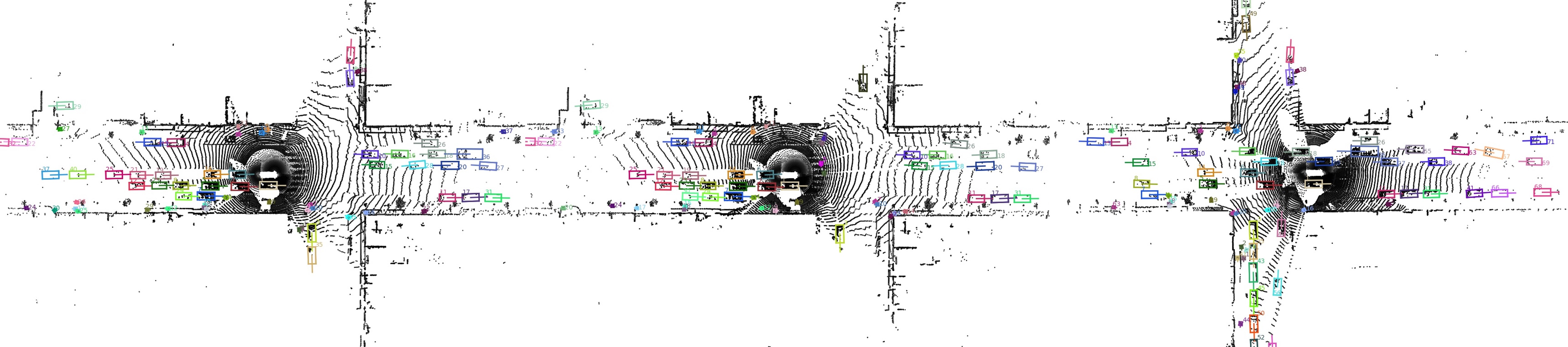}  
\label{fig:subfig2}
}\\
\subfloat{
\includegraphics[width=0.9\linewidth]{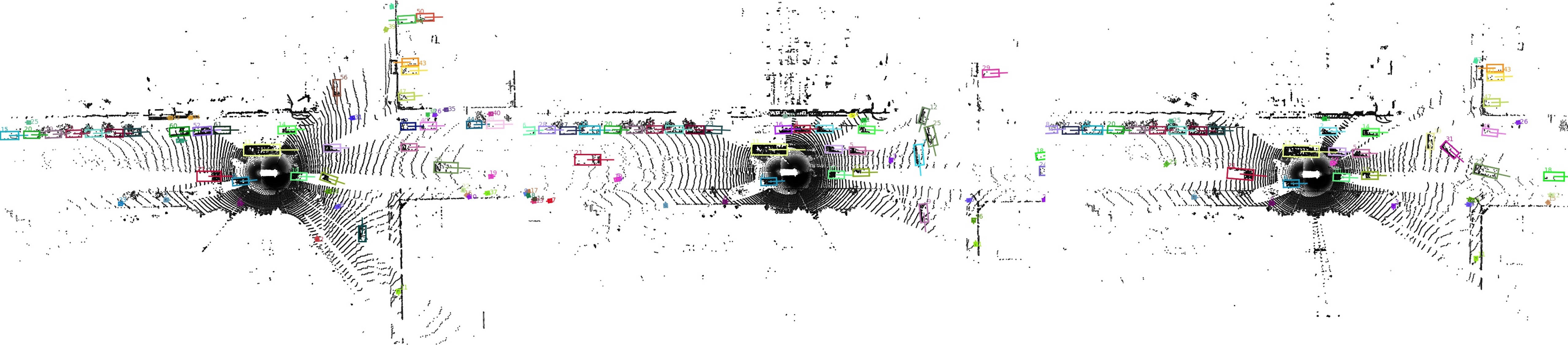}  
\label{fig:subfig3}
}\\
\subfloat{
\includegraphics[width=0.9\linewidth]{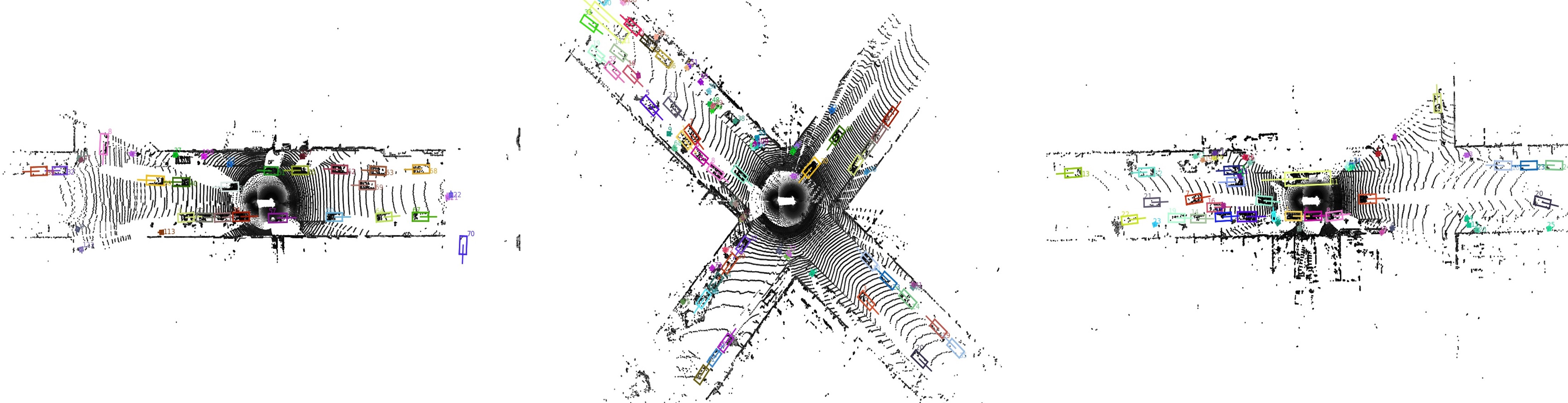}  
 \label{fig:subfig4}
}\\
\subfloat{
\includegraphics[width=0.9\linewidth]{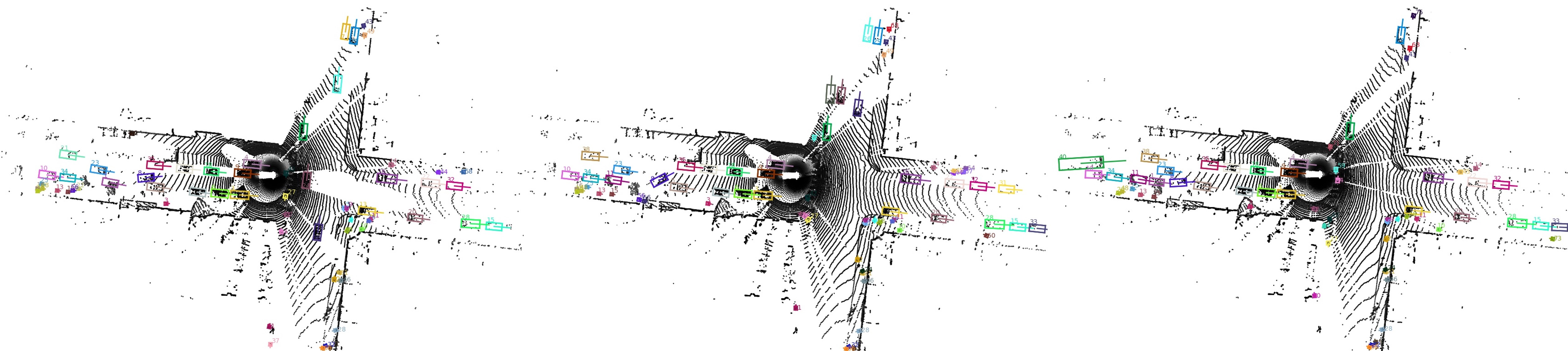}  
 \label{fig:subfig5}
}
\caption{Qualitative results of 3D tracking. Each row shows 3 frames from the same sequence. Each object is showed in a unique color and assigned a unique ID. Note that they are not consecutive frames, the time interval is larger than one frame. For better visualization purpose point cloud data is sub-sampled.}
\label{fig:3d_tracking_qualitative_results} 
\end{figure*}

As shown in Table~\ref{tab:2d_mot_results} and Table~\ref{tab:3d_mot_results}, our tracking algorithm reached the 1st place on the official Waymo Open Dataset 2D and 3D tracking leaderboards ~\cite{Waymo2DTracking2020} ~\cite{Waymo3DTracking2020} and achieved the highest MOTA$/$L2 scores. In particular, our trackers return the lowest miss rate. Some qualitative results are shown in Figure~\ref{fig:2d_tracking_qualitative_results} and Figure~\ref{fig:3d_tracking_qualitative_results}.

\subsection{Ablation Study}
On the 2D tracking validation set with 202 sequences, we study the effect of introducing the 3rd-stage association and using the Re-ID models. Our baseline performance is produced without these two components. As shown in Table~\ref{tab:2d_ablation_study}, the 3rd-stage association results in a $3.65\%$ improvement in terms of MOTA$/$L2 and the Re-ID models can further improve the performance by $2.43\%$.

\section{Conclusion}
An accurate, online and unified 2D and 3D tracking framework is proposed and achieved the 1st places on the Waymo Open Dataset 2D and 3D tracking challenges. In the future, we will continue our ongoing work with the above-mentioned joint detection and tracking framework.

{\small
\bibliographystyle{ieee_fullname}
\bibliography{egbib.bib}
}

\end{document}